\documentclass[runningheads]{llncs}
\usepackage{graphicx}
\usepackage{breakcites}
\usepackage{tikz}
\usepackage{comment}
\usepackage{amsmath,amssymb}  
\usepackage{graphicx}
\usepackage{stmaryrd}
\usepackage{booktabs}
\usepackage{breqn}
\usepackage{pifont}
\usepackage{makecell}
\usepackage{xcolor}
\usepackage{enumitem}
\usepackage{multirow}
\usepackage{wrapfig,blindtext} 
\usepackage{color, colortbl}
\usepackage{fancyhdr }
\usepackage{subcaption} 
\usepackage{url}
\usepackage[pagebackref,breaklinks,colorlinks,citecolor=blue]{hyperref}
\usepackage{lipsum} 
\usepackage[ruled,linesnumbered]{algorithm2e}   
\definecolor{Gray}{gray}{0.9} 
\usepackage[accsupp]{axessibility} 
\definecolor{darkorange}{rgb}{1.0, 0.55, 0.0} 
\DeclareUnicodeCharacter{3000}{  }
\DeclareUnicodeCharacter{2212}{-}
\newcommand{\btheta}{\boldsymbol{\theta}}
\newcommand{\bbeta}{\boldsymbol{\beta}}
\newcommand{\btau}{\boldsymbol{\tau}}
\newcommand{\bphi}{\boldsymbol{\phi}}
\newcommand{\bPhi}{\boldsymbol{\Phi}}

\newcommand{\bX}{\mathbf{X}}
\newcommand{\bI}{\mathbf{I}}
\newcommand{\bS}{\mathbf{S}}
\newcommand{\bV}{\mathbf{V}}
\newcommand{\bP}{\mathbf{P}}

\newcommand{\bp}{\mathbf{p}}
\newcommand{\bZ}{\mathbf{Z}}

\newcommand{\vf}{\mathbf{f}}
\newcommand{\vF}{\mathbf{F}}
\newcommand{\bc}{\mathbf{c}}
\newcommand{\bw}{\mathbf{w}}
\definecolor{better}{rgb}{0.19, 0.55, 0.91}
\newcommand{\cmark}{\textcolor{better}{\ding{51}}}
\newcommand{\xmark}{\textcolor{red}{\ding{55}}}
\begin{document} 
\pagestyle{headings}
\mainmatter
\title{HULC: 3D HUman Motion Capture with Pose \hbox{Manifold SampLing and Dense Contact Guidance}} 

\titlerunning{HULC: 3D HUman Motion Capture with Pose Manifold Sampling\ldots} 
%
\author{Soshi Shimada\inst{1} \;\;\;  Vladislav Golyanik\inst{1} \;\;\;  Zhi Li\inst{1}\;\;\; 
Patrick Pérez\inst{2}\;\;\; \\
Weipeng Xu\inst{1}\;\;\; 
Christian Theobalt\inst{1} 
}
\authorrunning{S. Shimada et al.}
%
\institute{ Max Planck Institute for Informatics, Saarland Informatics Campus \and
 Valeo.ai }
 
\maketitle

\begin{abstract}
Marker-less monocular 3D human motion capture (MoCap) with scene interactions is a challenging research topic relevant for extended reality, robotics and virtual avatar generation. 
Due to the inherent depth ambiguity of monocular settings, 3D motions captured with existing methods often contain severe artefacts such as incorrect body-scene inter-penetrations, jitter and body floating. 
To tackle these issues, we propose HULC, a new approach for 3D human MoCap which is aware of the scene geometry. 
HULC estimates 3D poses and dense body-environment surface contacts for improved 3D localisations, as well as the absolute scale of the subject.  
Furthermore, we introduce a 3D pose trajectory optimisation based on a novel pose manifold sampling that resolves erroneous body-environment inter-penetrations.  
Although the proposed method requires less structured inputs compared to existing scene-aware monocular MoCap algorithms, it produces more physically-plausible poses: HULC significantly and consistently outperforms the existing approaches in various experiments and on different metrics. Project page:  \url{https://vcai.mpi-inf.mpg.de/projects/HULC/}. 
\keywords{3D Human MoCap, dense contact estimations, sampling.}  
\end{abstract}

\section{Introduction}\label{sec:intro} 

3D human motion capture (MoCap) from a single colour camera received a lot of attention over the past years  \cite{VNect_SIGGRAPH2017,XNect_SIGGRAPH2020,hmrKanazawa17,humanMotionKanazawa19,kocabas2019vibe,pavllo20193d,choi2020beyond,tekin2016structured,mono-3dhp2017,rhodin2018unsupervised,chen_2017_3d,JMartinezICCV2017,tome2017lifting,Moreno-Noguer_2017,pavlakos2018learning,newell2016stacked,pavlakos2017volumetric,Yang_3dposeCVPR2018,Zhou_2017_ICCV,inthewild3d_2019,bogo2016smpl,wei2010videomocap,kolotouros2019spin,shi2020motionet,sun2019human,Kocabas_PARE_2021,Kocabas_SPEC_2021,kolotouros2021prohmr}. 
Its applications range from mixed and augmented reality, to movie production and game development, to immersive virtual communication and telepresence. 
MoCap techniques that not only focus on humans \textit{in a vacuum} but also account for the scene environment---this encompasses awareness of the physics or constraints due to the underlying scene geometry---%
are coming increasingly into focus 
\cite{PhysAwareTOG2021,PhysCapTOG2020,rempe2020contact,PROX:2019,Zhang:ICCV:2021,Zanfir2018,rempe2021humor,PIPCVPR2022}. 

Taking into account interactions between the human and the environment in MoCap poses many challenges, as not only  articulations and global translation of the subject must be accurate, but also contacts between the human and the scene need to be plausible. 
%
%
%
A misestimation of only a few parameters, such as a 3D translation, can lead to reconstruction artefacts that contradict physical reality (\textit{e.g.,} body-environment penetrations or body floating). 
\begin{wraptable} {t!}{0.53\textwidth} 
\center 
 \scalebox{0.50}{ 
\begin{tabular}{ l c  c c c c c c }\toprule 
 \multirow{3}{*}{Approach}  &  \multirow{3}{*}{Inputs} &      \multicolumn{5}{c }{Outputs}   \\  
 				  \cmidrule(lr){3-7} 
 &      &\shortstack{body\\pose} & $\btau$ & \shortstack{absolute \\ scale} &\shortstack{body\\contacts} &\shortstack{env. \\contacts}  \\  \addlinespace[2pt] \midrule
\rowcolor{Gray}
PROX \cite{PROX:2019} & RGB $\!+\!$ scene mesh & \cmark & \cmark & \xmark & \xmark & \xmark \\  
PROX-D \cite{PROX:2019}	& RGBD $\!+\!$ scene mesh & \cmark & \cmark & \xmark & \xmark & \xmark \\  
\rowcolor{Gray}
LEMO \cite{Zhang:ICCV:2021} & RGB(D) $\!+\!$ scene mesh &\cmark & \cmark & \xmark & \cmark$^*$ & \xmark\\  
HULC (ours)~ & RGB \!$+$\! scene point cloud   &\cmark & \cmark & \cmark & \cmark & \cmark   \\ \bottomrule  
\end{tabular}
}
\caption{\label{tab:in_out_com}  
 Overview of inputs and outputs of different methods.  
 ``$\btau$'' and ``env.~contacts'' denote global translation and environment contacts, respectively.
 $``*"$ stands for sparse marker contact labels.}
\end{wraptable} 
On the other hand, known human-scene contacts can serve as reliable boundary conditions for improved 3D pose estimation and localisation. 
While several algorithms merely consider human interactions with a ground plane  \cite{PhysAwareTOG2021,PhysCapTOG2020,rempe2020contact,rempe2021humor,Zanfir2018}, a few other methods also account for the contacts and interactions with the more general 3D environment \cite{PROX:2019,Zhang:ICCV:2021}. 
However, due to the depth ambiguity of the monocular setting, their estimated subject's root translations can be inaccurate, which can create implausible body-environment collisions.  
Next, they employ a body-environment collision  penalty as a soft constraint. 
Therefore, the convergence of the optimisation to a bad local minima can also cause unnatural body-environment collisions. 
This paper addresses the limitations of the current works and proposes a new 3D \textbf{HU}man MoCap framework with pose manifold  samp\textbf{L}ing and guidance by body-scene \textbf{C}ontacts,  abbreviated as HULC. It improves over other monocular 3D human MoCap methods that consider constraints from 3D scene priors \cite{PROX:2019,Zhang:ICCV:2021}. 
Unlike existing works, HULC estimates contacts not only on the human body surface but also on the environment surface for the improved global 3D translation estimations. 
Next, HULC introduces a pose manifold sampling-based optimisation to obtain plausible 3D poses while handling the severe  body-environment collisions in a \textit{hard manner}. 
Our approach regresses more accurate 3D motions respecting scene constraints while requiring less-structured inputs (\textit{i.e.,} an RGB image sequence and a point cloud of the static background scene) compared to the related monocular scene-aware methods \cite{PROX:2019,Zhang:ICCV:2021} that require a complete mesh and images.  
HULC returns physically-plausible motions, an absolute scale of the subject and dense contact labels both on a human template surface model and the environment. 
HULC features several innovations which in interplay enable its functionality, 
\textit{i.e.,} 1) a new learned implicit function-based dense contact label estimator for humans and the general 3D scene environment, and 2) a new pose optimiser for scene-aware pose estimation based on a pose manifold sampling policy. 
The first component allows us to jointly estimate the absolute subject's scale and its highly accurate root 3D translations.%
The second component prevents severe body-scene collisions and acts as a hard constraint, in contrast to widely-used soft collision losses \cite{PROX:2019,AMASS:ICCV:2019}. To train the dense contact estimation networks, we also annotate contact labels on a large scale synthetic daily motion dataset: GTA-IM \cite{caoHMP2020}. 
To summarise, our primary technical contributions are as follows: 
\begin{itemize}
\itemsep0em
\item A new 3D MoCap framework with simultaneous 3D human pose localisation and body scale estimation guided by estimated contacts. 
It is the first method that regresses the dense body and environment contact labels from an RGB sequence and a point cloud of the scene using an implicit function (Sec.~\ref{ssec:con_est_scene}).
\item A new pose optimisation approach with a novel pose manifold sampling yielding better results by imposing hard constraints on incorrect body-environment interactions (Sec.~\ref{ssec:pose_mani_opt_root}). 
\item Large-scale body contact annotations on the GTA-IM dataset \cite{caoHMP2020} that provides synthetic 3D human motions in a variety of scenes (Fig.~\ref{fig:overview_dataset} and Sec.~\ref{sec:datasets}). 
\end{itemize}

We report quantitative results,  
including an ablative study, which show that HULC outperforms existing methods in 3D accuracy and on physical plausibility metrics (Sec.~\ref{sec:evaluations}). 
See our video for qualitative comparisons. 
\section{Related Works}
\label{sec:related_works}

\smallskip\noindent\textbf{Most monocular MoCap approaches} %
estimate 3D poses alone or along with the body shape from an input image or video \cite{inthewild3d_2019,hmrKanazawa17,humanMotionKanazawa19,kocabas2019vibe,choi2020beyond,tekin2016structured,mono-3dhp2017,rhodin2018unsupervised,chen_2017_3d,jiang2020mpshape,JMartinezICCV2017,tome2017lifting,Moreno-Noguer_2017,pavlakos2018learning,newell2016stacked,pavlakos2017volumetric,Yang_3dposeCVPR2018,Zhou_2017_ICCV,inthewild3d_2019,bogo2016smpl,wei2010videomocap,kolotouros2019spin,shi2020motionet,sun2019human,Kocabas_PARE_2021,kolotouros2021prohmr,Zhang_2020_CVPR}. Some methods also estimate 3D translation of the subject in addition to the 3D poses  \cite{VNect_SIGGRAPH2017,XNect_SIGGRAPH2020,Kocabas_SPEC_2021,pavllo20193d}. 
Fieraru \textit{et al.}~\cite{Fieraru_2020_CVPR} propose a multi-person 3D  reconstruction method considering human-human interactions. 
Another algorithm class incorporates an explicit physics model into MoCap and avoids environmental collisions  \cite{PhysCapTOG2020,rempe2020contact,PhysAwareTOG2021,Yuan_2021_CVPR}. 
These methods consider interactions with only a flat ground plane or a stick-like object \cite{li2019motionforcesfromvideo}, 
unlike our HULC, that can work with arbitrary scene geometry. 

\smallskip\noindent\textbf{Awareness of human-scene contacts} is helpful for the estimation and synthesis \cite{wang2021synthesizing,hassan2021stochastic} of plausible 3D human motions. 
Some existing works regress sparse joint contacts on a kinematic skeleton  \cite{li2019motionforcesfromvideo,PhysAwareTOG2021,PhysCapTOG2020,rempe2020contact,rempe2021humor,Zou2020} or sparse markers \cite{Zhang:ICCV:2021}. 
A few approaches forecast contacts on a dense human mesh  surface \cite{Hassan:CVPR:2021,Mueller:CVPR:2021}. 
Hassan \textit{et al.}~\cite{Hassan:CVPR:2021} 
place a human in a 3D scene considering the semantic information and dense human body contact labels.
M\"uller \textit{et al}.~\cite{Mueller:CVPR:2021} propose a dataset with discrete annotations for self-contacts on the human body. 
Consequently, they apply a self-contact loss 
for more plausible final 3D poses. 
Unlike the existing works, our algorithm estimates vertex-wise dense contact labels on the  human body surface from an RGB input only. 
Along with that, it also regresses dense contact labels on the environment given the scene point cloud along with the RGB sequence. 
The simultaneous estimation of the body and scene contacts allows 
HULC to disambiguate the depth and scale of the subject, although only a single camera view and a single scene point cloud are used as inputs. 

\smallskip\noindent\textbf{Monocular MoCap with scene interactions.} 
Among the scene-aware MoCap approaches  \cite{PhysAwareTOG2021,PhysCapTOG2020,rempe2020contact,PROX:2019,Zhang:ICCV:2021,rempe2021humor,Zanfir2018}, there are a few ones that consider human-environment interactions given a highly detailed scene geometry \cite{PROX:2019,Zhang:ICCV:2021,MoCapDeform}. 
PROX (PROX-D)\cite{PROX:2019} estimates 3D motions given RGB (RGB-D) image, along with an input geometry provided as a signed distance field (SDF). 
Given an RGB(D) measurement and a mesh of the environment, LEMO \cite{Zhang:ICCV:2021} also produces geometry-aware global 3D human motions with an improved motion quality characterised by smoother transitions and robustness to occlusions thanks to the learned motion priors. 
%
%
These two algorithms require an RGB or RGB-D sequence with SDF (a 3D scan of the scene) or occlusion masks. 
In contrast, our HULC requires only an RGB image sequence and a point cloud of the scene; it returns dense contact labels on 1) the human body and 2) the environment, 3) global 3D human motion with translations and 4) absolute scale of the human body.  
See Table \ref{tab:in_out_com} for an overview of the characteristics.  
Compared to PROX and LEMO, HULC shows significantly-mitigated body-environment collisions.  

\smallskip\noindent\textbf{Sampling-based human pose tracking.} 
Several sampling-based human pose tracking algorithms were proposed. 
Some of them utilise particle-swarm optimisation \cite{john2010markerless,saini2013human,saini2012markerless}. 
Charles \textit{et al.}~\cite{AutomaticCharles2013} employ Parzen windows 
for 2D joints tracking. 
Similar to our HULC, Sharma~\textit{et al.} \cite{9008113} %
generate 3D pose samples by a conditional variational autoencoder (VAE)  \cite{sohn2015learning} conditioned on 2D poses. 
In contrast, we utilise the learned pose manifold of VAE for sampling, which helps to avoid local minima 
and prevent %
body-scene collisions. Also, unlike \cite{9008113}, we sample around a latent vector obtained from the VAE's encoder to obtain poses that are plausible and similar to the input 3D pose. 

\section{Method} 
 \label{sec:methods} 

\begin{figure}[t!]
\centerline{\includegraphics[width=0.98\linewidth]{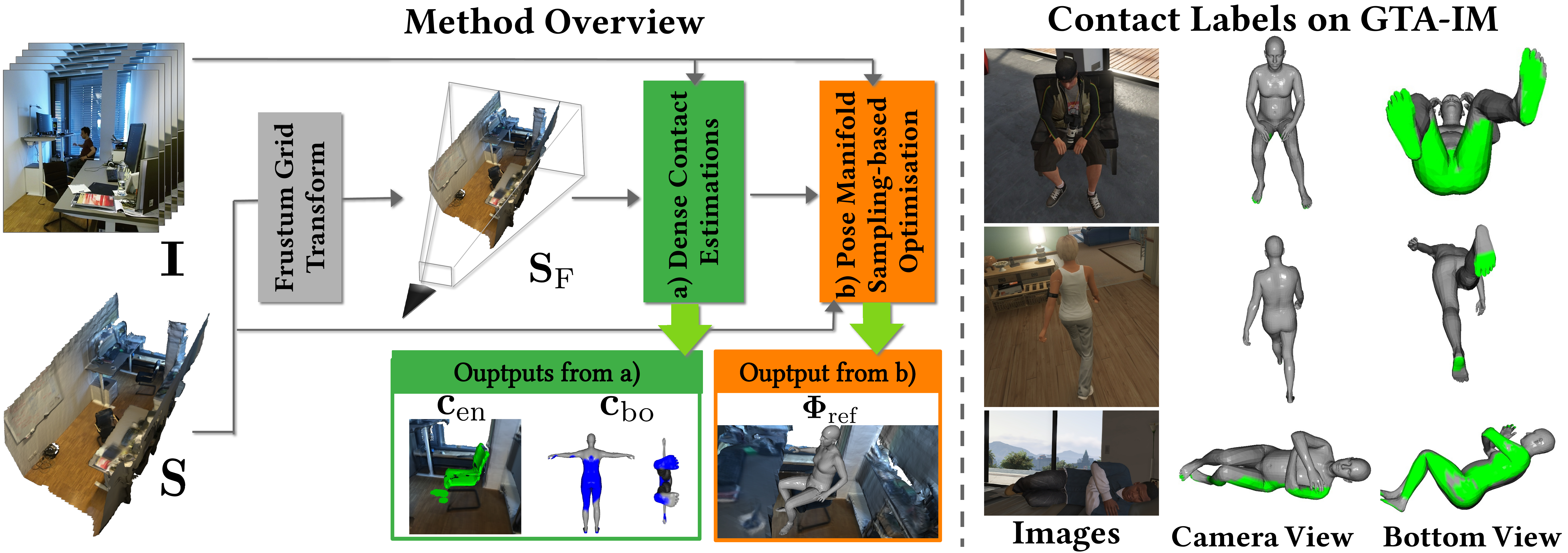}}
\caption{ 
(\textit{Left}) Given image sequence $\bI$, scene point cloud $\bS$ and its associated frustum voxel grid $\bS_{\text{F}}$, 
HULC first predicts for each frame dense contact labels on the body $\bc_{\text{bo}}$, 
and on the environment $\bc_{\text{en}}$.
It then refines initial, physically-inaccurate and scale-ambiguous global 3D poses $\bPhi_{0}$ into the final ones  $\bPhi_{\text{ref}}$ in (b).  Also see Fig.~\ref{fig:overview_a_b_uptoscale} for the details of stage (a) and (b). (\textit{Right}) Example visualisations of our contact annotations (shown in green) on GTA-IM dataset \cite{caoHMP2020}.
}
\label{fig:overview_dataset} 
\end{figure}

Given monocular video frames and a point cloud of the scene registered to the coordinate frame of the camera, our goal is to infer physically-plausible global 3D human poses along with dense contact labels on both body and environment surfaces. 
Our approach consists of two stages (Fig.\,\ref{fig:overview_dataset}):
\begin{itemize}\itemsep0em
    \item \textbf{Dense Body-environment contacts estimation}: Dense contact labels are predicted on body and scene surfaces using a learning-based approach with a pixel-aligned implicit representation inspired by \cite{saito2019pifu} (Sec.\,\ref{ssec:con_est_scene}); 
    \item \textbf{Sampling-based optimisation on the pose manifold}: We combine sampling in a learned latent pose space with gradient descent to obtain the absolute scale of the subject and its global 3D pose, under hard guidance by predicted contacts. 
    This approach significantly improves the accuracy of the estimated root translation and articulations, and mitigates incorrect environment penetrations.  (Sec.~\ref{ssec:pose_mani_opt_root}).
\end{itemize}

\paragraph{Modelling and Notations.} 
Our method takes as input a sequence $\bI\,{=}\,\{\bI_{1},...,\bI_{T}\}$ of $T$ successive video frames from a static camera with known intrinsics ($T\,{=}\,5$ in our experiments). 
We detect a squared bounding box around the subject and resize the cropped image region to $225\,{\times}\,225$ pixels. 
The background scene's geometry that corresponds to the detected bounding box is represented 
by a single static point cloud $\bS\,{\in}\,\mathbb{R}^{M\times 3}$ composed of $M$ points aligned in the camera reference frame in an absolute scale. 
To model the 3D pose and human body surface, we employ the parametric model SMPL-X \cite{SMPL-X:2019} (its gender-neutral version). 
This model defines the 3D body mesh as a differentiable function $\mathcal{M}(\btau, \bphi,\btheta,\bbeta)$ of 
global root translation $\btau\,{\in}\,\mathbb{R}^{3}$, global root orientation $\bphi\,{\in}\,\mathbb{R}^{3}$, root-relative pose $\btheta\,{\in}\,\mathbb{R}^{3K}$ of $K$ joints  
and shape parameters $\bbeta\,{\in}\,\mathbb{R}^{10}$ capturing body's identity. 
For efficiency, we downsample the original SMPL-X body mesh with over $10$k vertices to $\bV\,{\in}\,\mathbb{R}^{N\times 3}$, where $N\,{=}\,655$. 
In the following, we denote $\bV =  \mathcal{M}(\bPhi,\bbeta)$, where $\bPhi=(\btau,\bphi,\btheta )$ denotes the kinematic state of the human skeleton, from which the global positions $\bX\,{\in}\,\mathbb{R}^{K\times 3}$ of the $K\,{=}\,21$ joints can be derived. 

\subsection{Contact Estimation in the Scene}
\label{ssec:con_est_scene}

We now describe our learning-based approach for contact labels estimation on the human body and environment surfaces; see Fig.\,\ref{fig:overview_dataset}-a) for an overview of this stage. 
The approach takes $\bI$ and $\bS$ as inputs. 
It comprises three fully-convolutional feature extractors, $N_1$, $N_2$ and $N_3$, and two fully-connected layer-based contact prediction networks, $\Omega_{\text{bo}}$ and $\Omega_{\text{en}}$, for body and environment, respectively. 

\begin{figure}[t!]
\centerline{\includegraphics[width=1\linewidth]{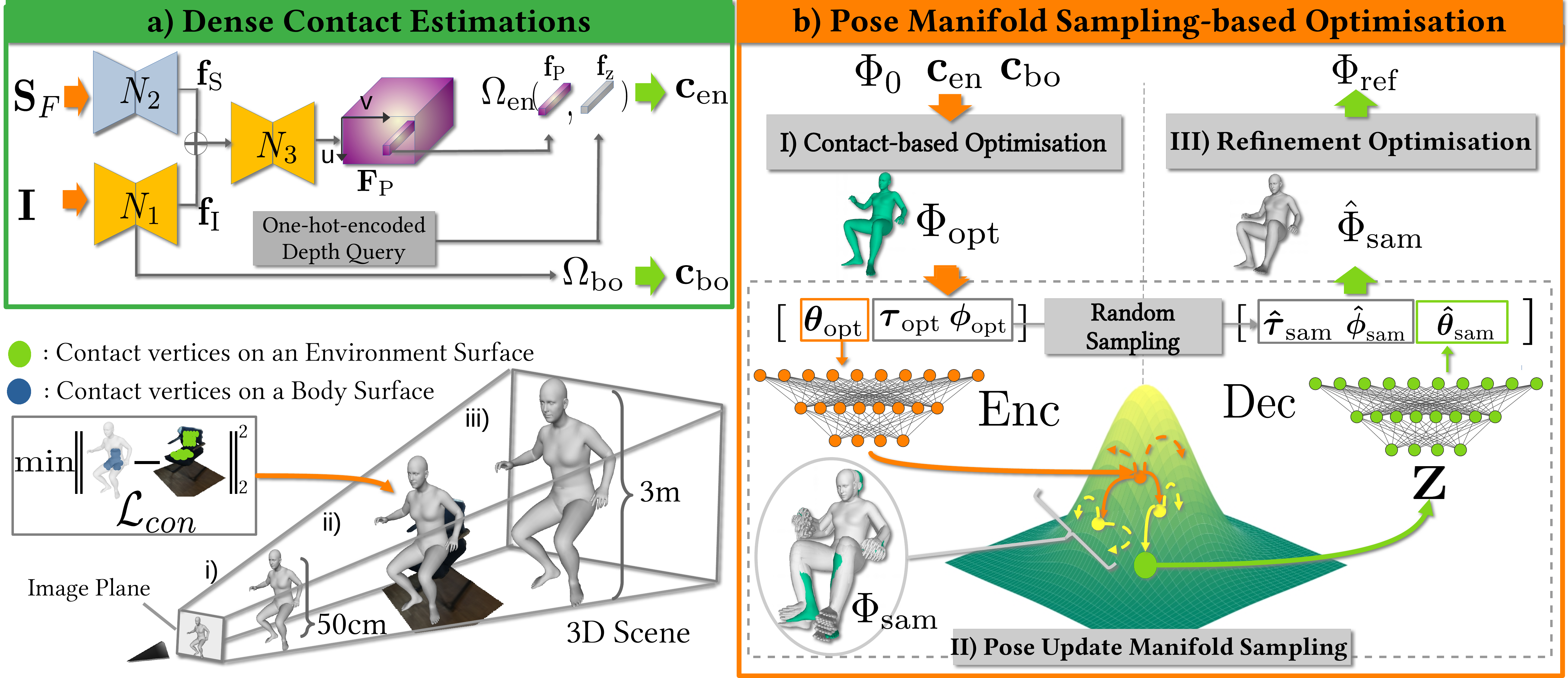}}
\caption{ \textbf{Overview of a) dense contact estimation and b) pose manifold sampling-based optimisation.}  
In b-II), we first generate samples around the mapping from $\boldsymbol{\theta}_{\text{opt}}$ (orange arrows), and elite samples are then selected among them (yellow points). 
After resampling around the elite samples (yellow arrows), the best sample is selected (green point).
The generated sample poses $\bPhi_{\text{sam}}$ (in gray color at the bottom left in b-II)) from the sampled latent vectors are plausible and similar to $\bPhi_{\text{opt}}$.  \textbf{(\textit{bottom left of the Figure})} Different body scale and depth combinations can be re-projected to the same image coordinates (i, ii and iii),\textit{i.e.}, \textbf{scale-depth ambiguity}. To simultaneously estimate the accurate body scale and depth of the subject (ii), we combine the body-environment contact surface distance loss $\mathcal{L}_{\text{con}}$ with the 2D reprojection loss.
}
\label{fig:overview_a_b_uptoscale} 
\end{figure} 

Network $N_{1}$ extracts from $\bI$ a stack of visual features $\vf_{\bI} \in \mathbb{R}^{32\times 32 \times 256}$. 
The latent space features of $N_1$ are also fed to $\Omega_{\text{bo}}$ to predict the vector $\bc_{\text{bo}}\in [0,1]^{N}$ of per-vertex contact probabilities on the \textit{body} surface. 

We also aim at estimating the  corresponding contacts on the \textit{environment} surface using an implicit function. 
To train a model that generalises well, 
we need to address two challenges: (i) No correspondence information between the scene points and the image pixels are given; (ii) Each scene contains a variable number of points. 
Accordingly, we convert the scene point cloud $\bS$ into a frustum voxel grid $\bS_{\text{F}} \in \mathbb{R}^{32\times 32 \times 256}$ (the third dimension corresponds to the discretised depth of the 3D space over 256 bins, please refer to our supplement for the details). 
This new representation is independent of the original point-cloud size and is aligned with the camera's view direction. 
The latter will allow us to leverage a pixel-aligned implicit function inspired by PIFu \cite{saito2019pifu}, which helps the networks figure out the correspondences between pixel and geometry information.  
More specifically, $\bS_{\text{F}}$ is fed into $N_{2}$, which returns scene features $\vf_{\text{S}}\in \mathbb{R}^{32\times 32\times 256}$. The third encoder, $N_{3}$, ingests $\vf_{\text{I}}$ and $\vf_{\text{S}}$ concatenated along their third dimension  and returns pixel-aligned features $\vF_{\text{P}}\in \mathbb{R}^{32\times 32\times 64}$. 
Based on $\vF_{\text{P}}$, $\Omega_{\text{en}}$ predicts the contact labels on the environment surface as follows. Given a 3D position in the scene, we extract the corresponding visual feature $\vf_{\text{P}} \in \mathbb{R}^{64} $ at the $(u, v)$-position in the image space from $\vF_{\text{P}}$ (via spacial bilinear interpolation), and query arbitrary depth with a one-hot vector $\vf_{\text{z}} \in \mathbb{R}^{256}$. 
We next estimate the contact labels  $c_{\text{en}}$ as follows: 
\begin{equation} \label{eq:pifu} 
   c_{\text{en}}=\Omega_{\text{en}}(\vf_{\text{P}},\vf_{\text{z}}). 
\end{equation}
Given contact ground truths $\hat{\bc}_{\text{bo}}\in\{0,1\}^N$ and $\hat{\bc}_{\text{en}}\in\{0,1\}^M$ on the body and the environment, the five networks are trained with the following loss: 
\begin{equation} \label{eq:con_train_loss}
   \mathcal{L}_{\text{labels}} =  
    \|\bc_{\text{en}} - \hat{\bc}_{\text{en}}\|^2_2
  + \lambda\,\operatorname{BCE}(\bc_{\text{bo}},\hat{\bc}_{\text{bo}}),
\end{equation} 
where $\operatorname{BCE}$ denotes the binary cross-entropy and  $\lambda = 0.3$. 
We use $\operatorname{BCE}$ for the body because the ground-truth contacts on its surface are binary; 
the $\ell_2$ loss is used for the environment, as sparse ground-truth contact labels are smoothed with a Gaussian kernel to obtain continuous signals. For further discussions of \eqref{eq:con_train_loss}, please refer to our supplement.
At test time, we only provide the 3D vertex positions of the environment 
to $\Omega_{\text{en}}(\cdot)$---to find the contact area on the scene point cloud---rather than all possible 3D sampling points as queries. 
This significantly accelerates the search of environmental contact labels while reducing the number of false-positive contact classifications. 
For more details of the network architecture, further discussions of the design choice and data  pre-processing, please refer to our supplement. 

\subsection{Pose Manifold Sampling-based Optimisation}
\label{ssec:pose_mani_opt_root}

In the second stage of the approach, we aim at recovering an accurate global 3D trajectory of the subject as observed in the video sequence, see Fig.~\ref{fig:overview_a_b_uptoscale}-(b) for the overview. An initial estimate $\bPhi_{0}$ is extracted for each input image using SMPLify-X \cite{SMPL-X:2019}.   
Its root translation $\btau$ being subject to scale ambiguity, we propose to estimate it more accurately, along with the actual scale $h$ of the person with respect to the original body model's height, under the guidance of the predicted body-environment contacts (\textbf{Contact-based Optimisation}).%
We then update the body trajectory and articulations in the scene, while mitigating the body-environment collisions with a new sampling-based optimisation on the pose manifold (\textbf{Sampling-based Trajectory Optimisation}).  
A subsequent refinement step yields the final global physically-plausible 3D  motions.

\smallskip\noindent\textbf{I) Contact-based Optimisation~}
Scale ambiguity is inherent to a monocular MoCap setting: 
Human bodies with different scale and depth combinations in 3D can be reprojected on the same positions in the image frame; see Fig.~\ref{fig:overview_a_b_uptoscale} and supplementary video for the schematic visualisation. 
Most existing algorithms that estimate global 3D translations of a subject either assume its known body scale 
\cite{PhysCapTOG2020,GraviCap2021,PhysAwareTOG2021} or use a statistical average body scale \cite{VNect_SIGGRAPH2017}.  
In the latter case, the estimated $\btau$ is often inaccurate and causes physically implausible body-environment penetrations.  
In contrast to the prior art, we simultaneously estimate 
$\btau$ and 
$h$ 
by making use of the body-environment dense contact labels from the previous stage (Sec.~\ref{ssec:con_est_scene}). 
For the given frame at time $t\!\in\!\llbracket1,T\rrbracket$, we select the surface regions with $\bc_{\text{en}}\,{>}\,0.5$ and $\bc_{\text{bo}}\,{>}\,0.5$ as effective contacts and leverage them in our optimisation. 
Let us denote the corresponding index subsets of body vertices and scene points by $\mathcal{C}_{\text{bo}}\,{\subset}\,\llbracket 1,N \rrbracket$ and  $\mathcal{C}_{\text{en}}\,{\subset}\,\llbracket 1,M \rrbracket$. 
The objective function for contact-based optimisation is defined as: 
\begin{equation} \label{eq:con_optim_all}
  \mathcal{L}_{\text{opt}}(\btau,h) = \lambda_{\text{2D}} \mathcal{L}_{\text{2D}}+  \lambda_{\text{smooth}}\mathcal{L}_{\text{smooth}}+\lambda_{\text{con}} \mathcal{L}_{\text{con}},
\end{equation} 
where the reprojection $\mathcal{L}_{\text{2D}}$, the temporal smoothness  $\mathcal{L}_{\text{smooth}}$ and the contact $\mathcal{L}_{\text{con}}$ losses weighted by empirically-set multipliers $\lambda_{\text{2D}}$, $\lambda_{\text{smooth}}$ and $\lambda_{\text{con}}$, read:
\begin{align}
\label{eq:reproj}
   \mathcal{L}_{\text{2D}}&=\frac{1}{K}\sum^{K}_{k=1} w_{k}\big\|\Pi(\bX_{k}) - \bp_{k}\big\|_2^{2}, \\
 \label{eq:smooth_trans}
  \mathcal{L}_{\text{smooth}}&= \left\|\btau - \btau_{\text{prev}} \right\|^{2}_{2},\\
 \label{eq:hausdorff}
   \mathcal{L}_{\text{con}}&= \sum_{n\in\mathcal{C}_{\text{bo}}} \min_{m\in\mathcal{C}_{\text{en}}}\left\|\bV_n-\bP_m\right\|^{2}_{2},
\end{align} 
where $\bp_k$ and $w_k$ are the 2D detection in the image of the $k$-th body joint and its associated confidence, respectively, obtained by OpenPose \cite{openpose4}; 
$\Pi$ is the perspective projection operator; 
$\btau_{\text{prev}}$ is the root translation estimated in the previous frame; $\bX_k$, $\bV_n$ and $\bP_m$ are, respectively, the $k$-th 3D joint, the $n$-th body vertex ($n \,{\in}\,  \mathcal{C}_{\text{bo}}$) and the $m$-th scene point ($m\,{\in}\, \mathcal{C}_{\text{en}}$). 
Note that the relative rotation and pose are taken from $\bPhi_{0}$. The body joints and vertices are obtained from  $\mathcal{M}$ using $\btau$ and  scaled with $h$. 
For $\mathcal{L}_{\text{con}}$, we use a directed Hausdorff measure \cite{HausdorffDist} as a distance 
between the body and environment contact surfaces.  
The combination of $\mathcal{L}_{\text{con}}$ and $\mathcal{L}_{\text{2D}}$ is key to disambiguate 
$\btau$ and 
$h$ (thus, resolving the monocular scale ambiguity). 
As a result of optimising \eqref{eq:con_optim_all} in frame $t$, we obtain $\bPhi_{\text{opt}}^t$, \textit{i.e.,} the global 3D human motion with absolute body scale. 
We solve jointly on $T$ frames and optimise for a single $h$ for them. 

\smallskip\noindent\textbf{II-a) Sampling-based Trajectory Optimisation~}
Although the poses $\bPhi_{\text{opt}}^t$, $t=1\cdots T$, estimated 
in the previous step  yield much more accurate 
$\btau$
and
$h$
compared to existing monocular RGB-based methods, incorrect body-environment penetrations are still observable. 
This is because the gradient-based optimisation often gets stuck in bad local minima (see the supplementary video for a toy example illustrating this issue). 
To overcome this problem, we introduce an additional sampling-based optimisation that imposes hard penetration constraints, thus significantly mitigating physically-implausible collisions. 
The overview of this algorithm is as follows: 
(i) For each frame $t$, we first draw candidate poses around $\bPhi_{\text{opt}}^t$ 
with a sampling function $\mathcal{G}$; 
(ii) The quality of these samples is ranked by a function $\mathcal{E}$ %
that allows selecting the most promising (``elite'') ones; samples with severe collisions are discarded; 
(iii) Using $\mathcal{G}$ and $\mathcal{E}$ again, we generate and select new samples around the elite ones. 
The details of these steps, $\mathcal{E}$ and $\mathcal{G}$, are elaborated next (dropping time index $t$ for simplicity). 

\smallskip\noindent\textbf{II-b) Generating Pose Samples.} 
We aim to generate $N_{\text{sam}}$ sample states $\bPhi_{\text{sam}}$ around the previously-estimated 
$\bPhi_{\text{opt}}=(\btau_{\text{opt}}, \bphi_{\text{opt}},\btheta_{\text{opt}})$. 
To generate samples $(\btau_{\text{sam}},\bphi_{\text{sam}})$ for the global translation and orientation, with 3DoF each, 
we simply use a uniform distribution around  $(\btau_{\text{opt}},\bphi_{\text{opt}})$; see our supplement for the details. 
However, na\"ively generating the relative pose $\btheta_{\text{sam}}$ in the same way around $\btheta_{\text{opt}}$ is highly inefficient because (i) the body pose is high-dimensional and (ii) the randomly-sampled poses are not necessarily plausible. 
These reasons lead to an infeasible amount of generated samples required to find a plausible collision-free pose; which is intractable on standard graphics hardware. 
To tackle these issues, we resort to the pose manifold learned by VPoser \cite{SMPL-X:2019}, which is a VAE \cite{kingma2013auto} trained on AMASS \cite{AMASS:ICCV:2019}, \textit{i.e.,} a dataset with many highly accurate MoCap sequences. Sampling is conducted in this VAE's latent space rather than in the kinematics pose space. 
Specifically, we first map $\btheta_{\text{opt}}$ into a latent pose vector %
with the VAE's encoder $\operatorname{Enc}(\cdot)$. 
Next, we sample latent vectors using a Gaussian distribution centered at this vector, with standard deviation $\boldsymbol{\sigma}$ (see Fig.\,\ref{fig:overview_a_b_uptoscale}-b). 
Each latent sample is then mapped through VAE's decoder $\operatorname{Dec}(\cdot)$ into a pose that is combined with the original one on a per-joint basis. 
The complete sampling process reads: 
\begin{equation}
    \bZ \sim \mathcal{N}\big(\operatorname{Enc}(\btheta_{\text{opt}}),\boldsymbol{\sigma}\big),\;\;
    \btheta_{\text{sam}} = \bw \circ \btheta_{\text{opt}} + (1-\bw) \circ \operatorname{Dec}(\bZ), 
    \label{eq:conf_merge}
\end{equation} 

where $\circ$ denotes Hadamard matrix product and
$\bw \,{\in}\,\mathbb{R}^{3K}$ is composed of the detection confidence values $w_{k}$, $k=1 \cdots K$, obtained from OpenPose, each appearing three times (for each DoF of the joint). 
This confidence-based strategy allows 
weighting higher the joint angles obtained by sampling,  
if the image-based detections are less confident (\textit{e.g.,} under occlusions). 
Conversely, significant modifications are not required for the joints with high confidence values. 

Since the manifold learned by VAE is smooth, the poses derived from the latent vectors sampled around $\operatorname{Enc}(\btheta_{\text{opt}})$ should be close to $\btheta_{\text{opt}}$. Therefore, we empirically set $\boldsymbol{\sigma}$ to a small value ($0.1$). 
Compared to the na\"ive random sampling in the joint angle space, whose generated poses are not necessarily plausible, this pose sampling on the learned manifold significantly narrows down the solution space. %
Hence, a lot fewer samples are required to escape  local minima. 
At the bottom left of Fig.~\ref{fig:overview_a_b_uptoscale}-b contains examples (gray color) of $\bPhi_{\text{sam}}$ ($N_{\text{sam}}\,{=}\,10$) overlayed onto $\bPhi_{\text{opt}}$ (green). 
In the following, we refer to this sample generation process as function $\mathcal{G}(\cdot)$. 

\smallskip\noindent\textbf{II-c) Sample Selection.} 
The quality of the $N_{\text{sam}}$ generated samples $\bPhi_{\text{sam}}$ is evaluated using the following cost function: 
\begin{align}
\label{eq:sampling_cost}
 \mathcal{L}_{\text{sam}}  &=  \mathcal{L}_{\text{opt}} +\lambda_{\text{sli}} \mathcal{L}_{\text{sli}}+\lambda_{\text{data}} \mathcal{L}_{\text{data}},\\
\label{eq:con_sliding}
  \mathcal{L}_{\text{sli}}&= \left\|\bV_{\text{c}}-\bV_{\text{c,pre}}  \right\|^{2}_{2},\\
\label{eq:ref_state}
  \mathcal{L}_{\text{data}}&= \left\|\bPhi_{\text{sam}} - \bPhi_{\text{opt}} \right\|^{2}_{2},
\end{align} 
where $\mathcal{L}_{\text{sli}}$ and $\mathcal{L}_{\text{data}}$ are contact sliding loss and data loss, respectively, and $\mathcal{L}_{\text{opt}}$ is the same as in \eqref{eq:con_optim_all} with the modification that the temporal consistency \eqref{eq:smooth_trans} applies to the whole $\bPhi_{\text{sam}}$; $\bV_{\text{c}}$ and $\bV_{\text{c,pre}}$ are the body contact vertices (with vertex indices in $\mathcal{C}_{\text{bo}}$) and their previous positions, respectively. 

Among $N_{\text{sam}}$ samples ordered according to their increasing $\mathcal{L}_{\text{sam}}$ values, 
the selection function $\mathcal{E}_{U}(\cdot)$ first discards those  causing stronger penetrations 
(in the sense that the amount of scene points inside a human body is above a threshold $\gamma$) 
and returns $U$ first samples from the remaining ones. 
If no samples pass the collision test, we regenerate the new set of $N_{\text{sam}}$ samples. 
This selection mechanism introduces the collision handling in a hard manner. 
After applying $\mathcal{E}_{U}(\cdot)$, with $U{<}N_{\text{sam}}$, $U$ elite samples are retained.  
Then, $\lfloor N_{\text{sam}}/U \rfloor$ new samples are regenerated around every elite sample using $\mathcal{G}$. 
Among those, the one with minimum $\mathcal{L}_{\text{sam}}$ value is retained as the final estimate. The sequence of obtained poses is temporally smoothed by Gaussian filtering to further remove jittering, which yields the global 3D motion $(\hat{\bPhi}^t_{\text{sam}}{} )_{t=1}^{T}$ with significantly mitigated collisions.

\smallskip\noindent\textbf{III) Final Refinement.}\label{sssec:refinement}  
From the previous step, we obtained the sequence $\hat{\bPhi}_{\text{sam}}=(\hat{\btau}_{\text{sam}}, \hat{\bphi}_{\text{sam}},\hat{\btheta}_{\text{sam}})$ of kinematic states whose severe body-environment collisions are prevented as hard constraints. Starting from these states as initialisation, we perform a final gradient-based refinement using cost function $\mathcal{L}_{\text{sam}}$ with $\hat{\bPhi}_{\text{sam}}$ replacing $\bPhi_{\text{opt}}$. The final sequence is denoted $(\bPhi^t_{\text{ref}})_{t=1}^{T}$. 

\section{Datasets with Contact Annotations}
\label{sec:datasets} 

As there are no publicly-available large-scale datasets with images and corresponding human-scene contact annotations, 
we annotate %
several existing datasets. 

\noindent\textbf{GTA-IM}~\cite{caoHMP2020} dataset contains various daily 3D motions. 
First, we fit SMPL-X model onto the 3D joint trajectories in GTA-IM. 
For each frame, we select contact vertices on the human mesh if: i) The Euclidean distance between the human body vertices on and the scene vertices are smaller than a certain threshold; ii) The velocity of the vertex is lower than a certain threshold. 
In total, we obtain the body surface contact annotations on $320k$ frames, which will be released for research purposes, see Fig.~\ref{fig:overview_dataset} for the examples of the annotated contact labels.

\noindent\textbf{PROX dataset} \cite{PROX:2019} contains scanned scene meshes, scene SDFs, RGB-D sequences, 3D human poses and shapes generated by fitting SMPL-X model onto the RGB-D sequences  (considering collisions). 
We consider the body vertices, whose SDF values are lower than $5$\,cm, as contacts. 
We annotate the environment contacts by finding the vertices that are the nearest to the body contacts.

\noindent\textbf{GPA dataset} \cite{gpa2019,cdg2020} contains multi-view image sequences of people interacting with various rigid 3D geometries, 
accurately reconstructed 3D scenes and 3D human motions obtained from VICON system \cite{vicon} with $28$ calibrated cameras. 
We fit SMPL-X on GPA to obtain the 3D shapes and compute the scene's SDFs to run other methods \cite{PROX:2019,Zhang:ICCV:2021,Hassan:CVPR:2021}.  

We extract from \textbf{GPA} $14$ test sequences with $5$ different subjects. 
We also split \textbf{PROX} \cite{PROX:2019} into training and test sequences. 
The training sequences of \textbf{PROX} and \textbf{GTA-IM} \cite{caoHMP2020} are used to train the contact estimation networks. For further details of dataset and training, please refer to our supplement.

\section{Evaluations}\label{sec:evaluations}

We compare our HULC with the most related scene-aware 3D MoCap algorithms, \textit{i.e.,}  PROX\cite{PROX:2019}, PROX-D\cite{PROX:2019}, POSA\cite{Hassan:CVPR:2021} and LEMO \cite{Zhang:ICCV:2021}. We also test SMPLify-X \cite{SMPL-X:2019} which does not use scene constraints. 
The root translation of SMPLify-X is obtained from its estimated camera poses as done in \cite{PROX:2019}. 
To run LEMO \cite{Zhang:ICCV:2021} on the RGB sequence, we use SMPLify-X\cite{SMPL-X:2019} to initialise it; we call this combination ``LEMO (RGB)''. 
\begin{table}[t!] \caption{\label{tab:error_3D} Comparisons of 3D error on GPA dataset \cite{gpa2019,cdg2020}. ``$\dagger$'' denotes that the occlusion masks for LEMO(RGB) were computed from GT 3D human mesh.}
\centering \scalebox{0.87}{\renewcommand{\arraystretch}{0.95}
\begin{tabular}{c c c c c c c }\toprule 
& \multicolumn{3}{c}{No Procrustes}& \multicolumn{3}{c}{Procrustes}\\
\cmidrule(lr){2-4}\cmidrule(lr){5-7}
&  MPJPE [mm]$\downarrow$ & PCK [$\%$]$\uparrow$ & PVE [mm]$\downarrow$ &  MPJPE [mm]$\downarrow$ & PCK [$\%$]$\uparrow$& PVE [mm]$\downarrow$ \\  
\addlinespace[2pt] \midrule
 Ours              &\bf{217.9}& \bf{35.3} &  \bf{214.7} &  \bf{81.5} &  \bf{89.3}  &\bf{72.6}\\ 
  Ours (w/o S)     & 221.3    &  34.5  &  217.2 &   82.6 &  \bf{89.3}  & 73.1\\ 
  Ours (w/o R)     &  240.8   &   31.9& 237.3     & 83.1 &    86.6     &  73.6\\ 
  Ours (w/o SR)    & 251.1    &  31.5 &  245.2  &   83.9 &  86.6   & 74.1 \\\Xhline{0.1pt}
  SMPLify-X \cite{SMPL-X:2019}  &  550.0   & 10.0 & 549.1 &   84.7     &  85.9 &   74.1 \\ 
  PROX \cite{PROX:2019}  &   549.7  & 10.1 & 548.7  &   84.6     &  86.0       &  73.9 \\ 
  POSA \cite{Hassan:CVPR:2021}  &  552.2   & 10.1  & 550.9  &   85.5     &  85.6   &  74.5 \\  
  LEMO (RGB)	\cite{Zhang:ICCV:2021}  &   570.1  & 8.75 & 570.5   &   83.0     &  86.4   &   73.7 \\
  LEMO (RGB) \cite{Zhang:ICCV:2021}$\dagger$ & 570.0  & 8.77    & 570.4  &   83.0     &  86.4       &  73.6\\   \bottomrule 
 \end{tabular}}  
\end{table}
\begin{table}[t!] 
\begin{minipage}{.44\textwidth}\caption{\label{tab:global_trans_Comparison} Ablations and comparisons for global translations and absolute body length on GPA dataset.}
\centering
      \scalebox{0.85}{\renewcommand{\arraystretch}{0.9}
      \begin{tabular} { ccc  }\toprule
       & \makecell{global \\ translation\\ error [m] $\downarrow$} & \makecell{absolute\\  bone length\\   error [m]$\downarrow$} \\   \midrule
     Ours (+1m) & \bf{0.242}& 0.104 \\
     Ours (+3m)  & 0.244 & \bf{0.097} \\
     Ours (+10m)  &  0.244& 0.109 \\\hline
     Baseline (+1m)  &  0.751 & 0.498  \\ 
     Baseline (+3m)  & 1.033 &  0.560 \\  
     Baseline (+10m)  & 2.861  &  1.918  \\ \hline
     SMPLify-X \cite{SMPL-X:2019}  &  0.527& 0.156 \\   
     PROX \cite{PROX:2019}   & 0.528& 0.160 \\ 
     POSA \cite{Hassan:CVPR:2021}  & 0.545 & 0.136 \\\bottomrule
\end{tabular}} 
\end{minipage} 
\begin{minipage}{.53\textwidth} \caption{\label{tab:plausibility_errors}Comparisons of physical plausibility measures on GPA dataset \cite{gpa2019,cdg2020} and PROX dataset  \cite{PROX:2019}.}
\centering
\scalebox{0.68}{ 
\renewcommand{\arraystretch}{1.05}
\begin{tabular}{ c c c  c  c  c c }\toprule 
&  &\multicolumn{2}{c}{GPA Dataset} & &\multicolumn{1}{c }{PROX Dataset}\\ 
\cmidrule(lr){3-4} \cmidrule(lr){5-6}
 &  & non penet. [$\%$]$\uparrow$ & $e_{\text{smooth}} \downarrow$ &  & non penet. [$\%$]$\uparrow$ \\
 \addlinespace[2pt] \midrule 
 \multirow{8}{*}{\rotatebox[origin=c]{0}{RGB}} 
&Ours   & \bf{99.4} & 20.2 && 97.0\\
&Ours (w/o S) & 97.6 &  28.1 && 93.8\\
&Ours (w/o R) &  \bf{99.4} & 24.7 && \bf{97.1} \\
&Ours (w/o SR) &  97.6   &  47.1 && 93.8\\ 
\cmidrule(lr){2-6} 
 &SMPLify-X \cite{SMPL-X:2019} & 97.7 & 43.3 && 88.9　\\ 
 & PROX \cite{PROX:2019} & 97.7 & 43.2 && 89.8 \\  
 & LEMO (RGB)\cite{Zhang:ICCV:2021} & 97.8 & \bf{19.9} && -\\   
 & POSA \cite{Hassan:CVPR:2021} & 98.0 &  47.0 && 93.0 \\ \midrule 
 \multirow{2}{*}{\rotatebox[origin=c]{0}{RGB-D}} 
 & PROX-D \cite{PROX:2019} & - &  - && 94.2\\  
 & LEMO \cite{Zhang:ICCV:2021} & - & - && 96.4\\ \bottomrule
\end{tabular}
}
\end{minipage}
\end{table}
We use the selected test sequences of GPA \cite{gpa2019,cdg2020} and PROX \cite{PROX:2019} dataset for the quantitative and qualitative comparisons. 
To avoid redundancy, we downsample all the predictions to $10$ fps except for the temporal consistency measurement ($e_{\text{smooth}}$ in Table \ref{tab:plausibility_errors}). Since the 3D poses in PROX dataset 
are prone to inaccuracies due to their human model fitting onto the RGB-D sequence, we use it only for reporting the body-scene penetrations (Table \ref{tab:plausibility_errors}) and for qualitative comparisons. 

\subsection{Quantitative Results}
\label{ssec:quantitative_results}
\begin{wrapfigure} [19]{R}{0.48\textwidth} 
	\begin{center} 
		\includegraphics[width=\linewidth]{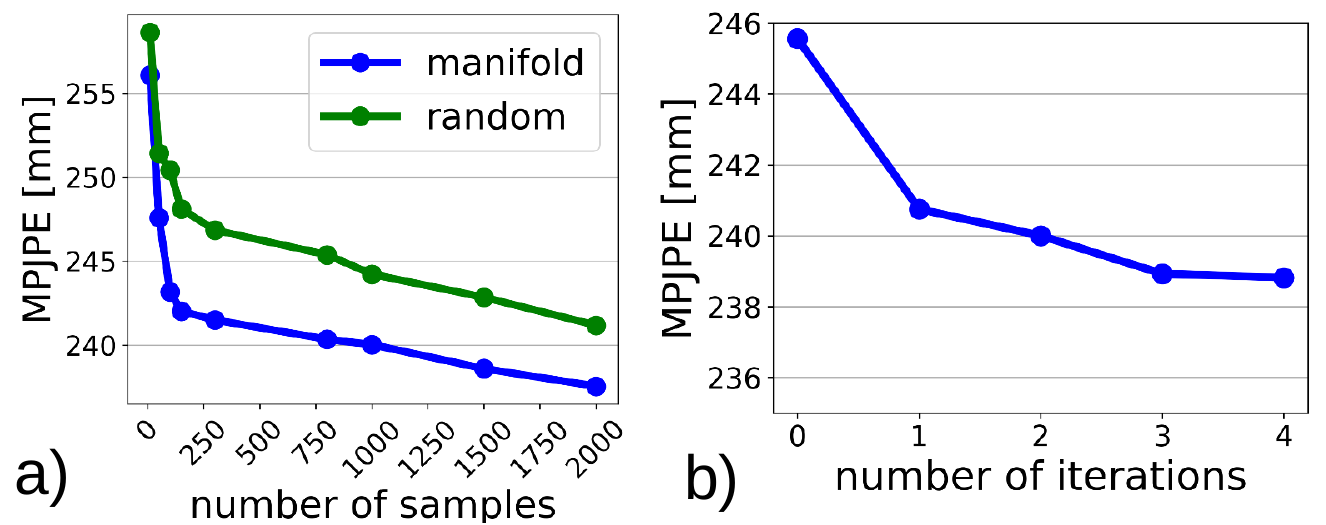} 
	\end{center} 
	\caption{(a) MPJPE [mm] comparison with different numbers of samples for the learned manifold sampling strategy vs. the na\"ive random sampling in the joint angle space of the kinematic skeleton.(b)  MPJPE [mm] comparison with different numbers of iterations in the sampling strategy.
	} \label{fig:ablations}	 
\end{wrapfigure}

We report 3D joint and vertex errors (Table \ref{tab:error_3D}), global translation and body scale estimation errors (Table \ref{tab:global_trans_Comparison}),  body-environment penetration and smoothness errors (Table \ref{tab:plausibility_errors}) and ablations on the sampling-based optimisation component, \textit{i.e.}, a) Manifold sampling vs. random sampling and b) Different number of sampling iterations in Fig.~\ref{fig:ablations}. 
``Ours (w/o S)'' represents our method without the sampling optimisation component, \textit{i.e.,} only the contact-based optimisation and refinement are applied (see Fig.~\ref{fig:overview_a_b_uptoscale}-(b) and Sec.~\ref{ssec:pose_mani_opt_root}). 
``Ours (w/o R)'' represents our method without the final refinement. 
``Ours (w/o SR)'' denotes ours without the sampling and refinement. 
For a further ablation study and evaluation of contact label estimation networks, please see our supplement.

\noindent\textbf{3D Joint and Vertex Errors.} Table \ref{tab:error_3D} compares the accuracy of 3D joint and vertex positions with and without Procrustes alignment.  
LEMO also requires human body occlusion masks on each frame.  
We compute them using the scene geometry and SMPLify-X  \cite{SMPL-X:2019} results. 
We also show another variant ``LEMO (RGB)$\dagger$'' whose occlusion masks are computed using the ground-truth global 3D human mesh instead of SMPLify-X.
Here, we report the standard 3D metrics, \textit{i.e.,} mean per joint position error (MPJPE), percentage of correct keypoints (PCK) (@150mm) and mean per vertex error (PVE). Lower MPJPE and PVE represent more accurate 3D reconstructions, higher PCK indicates more accurate 3D joint positions.
On all these metrics, HULC outperforms other methods both with and without Procrustes. 
Notably, thanks to substantially more accurate global translations obtained from the contact-based optimisation (Sec.~\ref{ssec:pose_mani_opt_root}), %
HULC  significantly reduces the MPJPE and PVE with a big margin, \textit{i.e.,} ${\approx}\,60\%$ error deduction in MPJPE and PVE w/o Procrustes compared to the second-best method. 
The ablative studies on Table \ref{tab:error_3D} also indicate that both the sampling and refinement optimisations contribute to accurate 3D poses. Note that the sampling optimisation alone (``Ours (w/o R)'') does not significantly reduce the error compared to ``Ours (w/o SR)''. This is because the sampling component prioritises removal of environment penetrations by introducing hard collision handling, which is the most important feature of this component. Therefore, the sampling component significantly contributes to reducing the environment collision as can be seen in Table \ref{tab:plausibility_errors} (discussed in the later paragraph). 
Applying the refinement after escaping from severe penetrations by the sampling optimisation further increases the 3D accuracy (``Ours'' in Table \ref{tab:error_3D}) while significantly mitigating physically implausible body-environment penetrations (Table \ref{tab:plausibility_errors}).

\noindent\textbf{Global Translation and Body Scale Estimation.} Table \ref{tab:global_trans_Comparison} reports global translation and body scale estimation errors for the ablation study of the contact-based optimisation (Sec.~\ref{ssec:pose_mani_opt_root}). %
More specifically, we evaluate the output $\Phi_{\text{opt}}$ obtained from the contact-based optimisation denoted ``ours''. 
We also show the optimisation result without using the contact loss term \eqref{eq:hausdorff} (``Baseline''). 
The numbers next to the method names represent the initialisation offset from the ground-truth 3D translation position (\textit{e.g.,} ``+$10$m'' indicates that the initial root position of the human body was placed at $10$ meters away along the depth direction from the ground-truth root position when solving the optimisations).
Without the contact loss term---since global translation and body scale are jointly estimated in the optimisation---the baseline method suffers from \textit{up-to-scale} issue (see Fig.~\ref{fig:overview_a_b_uptoscale}). 
Hence, its results are significantly worse due to worse initialisations. 
In contrast, our contact-based optimisation disambiguates %
the scale and depth by localising the contact positions on the environment, which confirms HULC to be highly robust to bad initialisations. 
Compared to the RGB-based methods PROX, POSA and SMPLify-X, our contact-based optimisation result has ${\approx}\,40\%$ smaller  error in the absolute bone length, and ${\approx}\,57\%$ smaller error in global translation, which also contributes to the reduced body-environment collisions as demonstrated in Table \ref{tab:plausibility_errors} (discussed in the next paragraph). 

\noindent\textbf{Plausibility Measurements.} 
We also report the plausibility of the reconstructed 3D motions in Table \ref{tab:plausibility_errors}. 
\textit{Non penet.} measures the average ratio of non-penetrating body vertices into the environment over all frames. 
A higher value denotes fewer body-environment collisions in the sequence; 
$e_{\text{smooth}}$ measures the temporal smoothness error proposed in \cite{PhysCapTOG2020}. 
Lower $e_{\text{smooth}}$ indicates more temporally smooth 3D motions. 
On both GPA and PROX datasets, our full framework mitigates the collisions thanks to the manifold sampling-based optimisations (ours vs. ours (w/o S)). 
It also does so 
when compared to other related works as well. 
Notably, HULC shows the least amount of collisions even  compared with RGBD-based methods on the PROX dataset. 
Finally, the proposed method also shows the significantly low $e_{\text{smooth}}$ (on par with LEMO(RGB)) in this experiment.

\noindent\textbf{More Ablations on Sampling-based Optimisation.} 
In addition to the ablation studies reported in Tables \ref{tab:error_3D}, \ref{tab:global_trans_Comparison} and \ref{tab:plausibility_errors}, we further assess the performance of the pose update manifold sampling step (Fig.~\ref{fig:overview_a_b_uptoscale}-(b)-(II)) on GPA dataset \cite{gpa2019,cdg2020}, reporting the 3D error (MPJPE [mm]) measured in world frame. Note that we report MPJPE without the final refinement step to assess the importance of the manifold sampling approach. In Fig.~\ref{fig:ablations}-(a), we show 
the influence of the number $N_{\text{sam}}$ of samples on 
 the performance of our manifold sampling strategy vs. a na\"ive random sampling with a uniform distribution in a kinematic skeleton frame. For the details of the na\"ive random sampling strategy, please refer to our supplement.
 In Fig.~\ref{fig:ablations}-(a), since the generated samples of the learned manifold return plausible pose samples, our pose manifold sampling strategy requires significantly fewer samples compared to the random sampling (${\sim}15\times$ more samples are required for the random sampling to reach 243 [mm] error in MPJPE). This result strongly supports the importance of the learned manifold sampling. No more than $2000$ samples can be generated due to the hardware memory capacity.
 In Fig.~\ref{fig:ablations}-(b), we report the influence of the number of generation-selection steps using functions $\mathcal{G}$ and $\mathcal{E}_{U}$ (with $U\,{=}\,3$) introduced in Section \ref{ssec:pose_mani_opt_root}, with $N_{\text{sam}}\,{=}\,1000$ samples. 
No iteration stands for choosing the best sample from the first generated batch (hence no resampling), while one iteration is the variant described in Sec. \ref{ssec:pose_mani_opt_root}. This first iteration sharply reduces the MPJPE, while the benefit of the additional iterations is less pronounced. 
Based on these observations, we use only one re-sampling iteration with $1000$ samples in the previous experiments.
Finally, we ablate the confidence value-based pose merging in Eq.\,(\ref{eq:conf_merge}), setting $N_{\text{sam}}{=}1000$ and the number of iterations to $0$. The measured MPJPE for with and without this confidence merging are $245.5$ and $249.1$, respectively.
\subsection{Qualitative Results}
\label{ssec:qualitative_results}
  \begin{figure*}[t]
	\begin{center}
		\includegraphics[width=\linewidth ]{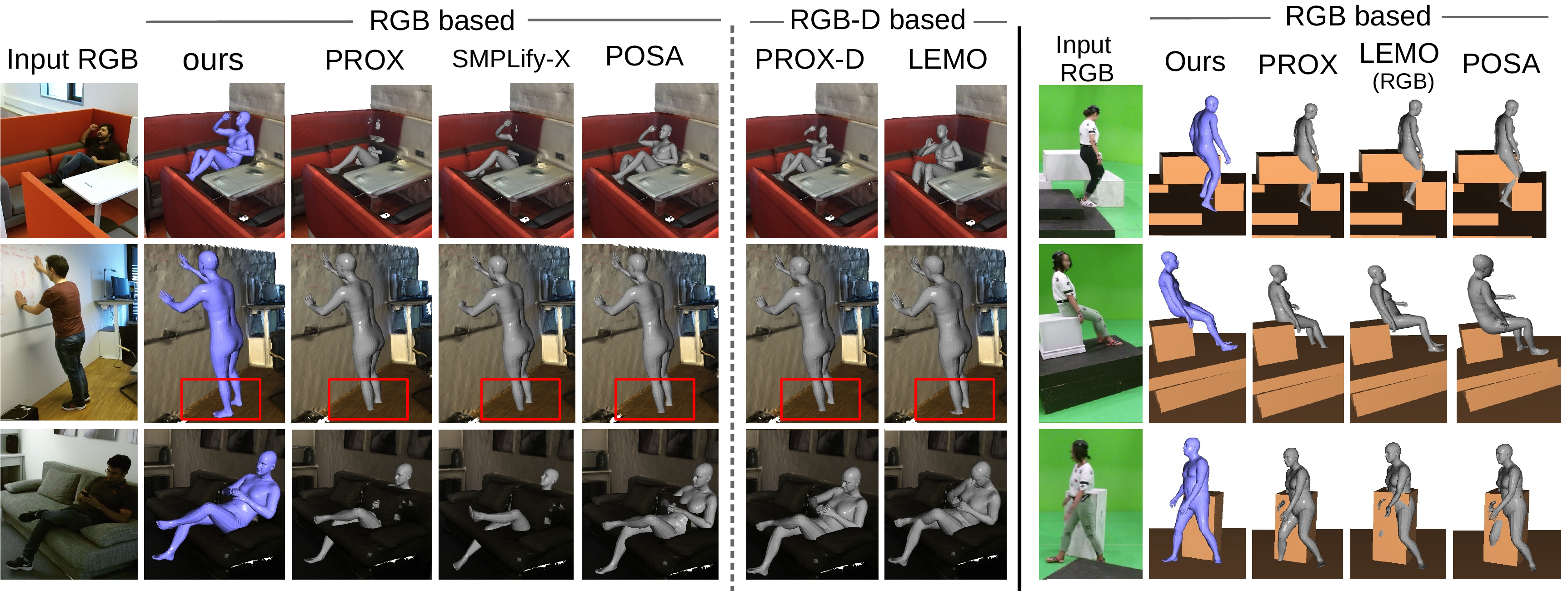}
	\end{center}
	\caption
	{ \label{fig:qual} The qualitative comparisons of our results with the related methods on PROX (left) and GPA dataset (right). Our RGB-based HULC shows fewer body-scene penetrations even when compared with RGB-D based methods; mind the red rectangles in the second row. 
	}  
\end{figure*}
 Figure~\ref{fig:qual} summarises the qualitative comparisons on GPA and PROX datasets. 
 HULC produces more physically-plausible global 3D poses with mitigated collisions, whereas the other methods show body-environment penetrations. 
 Even compared with the RGB(D) approaches, HULC mitigates collisions (mind the red rectangles). For more qualitative results, please refer to our video.

\section{Concluding Remarks}
\label{sec:conclusions}
\noindent{\textbf{Limitations.}} HULC requires the scene geometry aligned in a camera frame like other related works \cite{PROX:2019,Zhang:ICCV:2021,Hassan:CVPR:2021}. Also, HULC does not capture non-rigid deformations of scenes and bodies, although the body surface and some objects in the environment deform (\textit{e.g.,} when sitting on a couch or lying in a bed).  
Moreover, since our algorithm relies on the initial root-relative pose obtained from an RGB-based MoCap algorithm, the subsequent steps can fail under severe occlusions. 
Although the estimated contact labels help to significantly reduce the 3D translation error, the estimated environment contacts contain observable false positives. 
These limitations can be tackled in the future. 

\noindent{\textbf{Conclusion.}} We introduced \textit{HULC}---the first RGB-based scene-aware MoCap algorithm that estimates and is guided by dense body-environment surface contact labels combined with a pose manifold sampling.  
HULC shows $60\%$ smaller 3D-localisation errors compared to the previous methods.  
Furthermore, deep body-environment collisions are handled in \text{hard manner} in the pose manifold sampling-based optimisation, which significantly mitigates collisions with the scene. 
HULC shows the lowest collisions even compared with RGBD-based scene-aware methods. 

\noindent{\textbf{Acknowledgements.}}The authors from MPII were supported by the ERC Consolidator Grant 4DRepLy (770784). We also acknowledge support from Valeo.

%
%
\bibliographystyle{splncs04}
\bibliography{egbib}
\appendix 
\clearpage
\noindent{\fontsize{12}{12}\selectfont \textbf{Appendices}}
\vspace{0.1cm}

This supplementary document provides further details on our framework (Sec.~\ref{sec:frame_work}), dataset and training details (Sec.~\ref{sec:data_train}), the details of contact classification networks with discussions of the architecture and loss function design choices (Sec.~\ref{sec:net_arch}), and additional evaluation for the dense contact estimation network and ablation studies for the optimisations in our framework, as well as the details of the qualitative comparisons (Sec.~\ref{sec:eva_and_abl}).

\section{Framework Details}
\label{sec:frame_work}
In this section, we elaborate on the details of the frustum grid transform (Sec.~\ref{ssec:frustum_conversion}), the implementation details (Sec.~\ref{ssec:implementations}), the optimisation details (Sec.~\ref{ssec:optim_details}) and the random sampling for the root translation and orientation in the sampling-based optimisation stage (Sec.~\ref{ssec:rand_samp}). We also explain, in Sec.~\ref{ssec:rand_samp_pose}, the details of the random sampling in the kinematic pose space used in the ablation study in the main paper (Sec. 5.1).
 
\subsection{Frustum Grid Transform}
\label{ssec:frustum_conversion}
We conduct as follows the transformation from the scene point cloud $\mathbf{S}\in \mathbb{R}^{M\times 3}$, defined in the camera frame, into the frustum voxel grid $\mathbf{S}_{\text{F}}\in \mathbb{R}^{32 \times 32 \times  256}$ whose third dimension corresponds to the discretised depth of the 3D space. 
Given a vertex position $p =(x,y,z)$, \textit{i.e.}, a row of $ \mathbf{S}$, in a perspective frustum space, its normalised vertex $\hat{p}$ into the cuboid space reads:
\begin{equation}
\label{eq:state}
\hat{p} =\left(f_{x}\frac{x}{z},f_{y}\frac{y}{z},{z}\right),
\end{equation}
where $f=(f_{x}, f_{y})$ is the camera's focal length.  
The components of all points $\hat{p}$ are then suitably normalised and binned so as to build the binary occupancy grid $\bS_{\text{F}}$. 
\subsection{Implementations}
\label{ssec:implementations}
The neural networks are implemented with PyTorch \cite{NEURIPS2019_9015} and Python 3.7.  We conducted the evaluations on a computer with one AMD EPYC 7502P 32 Core Processor and one NVIDIA QUADRO RTX 8000 graphics card. The training of the contact classification networks continued until the loss convergence using Adam optimiser \cite{adam} with a learning rate $3.0 \times 10^{−4}$. Our framework runs with 25 seconds per frame excepting the computation time of SMPLify-X \cite{SMPL-X:2019} which we use for the initial root-relative pose estimation.
\subsection{Optimisation Details}
\label{ssec:optim_details}
For the optimization in Eq.\,3 of the main paper, we use the weights $\lambda_{\text{2D}}=1.0$, $\lambda_{\text{smooth}}=0.01$ and $\lambda_{\text{con}}=0.01$. For Eq.\,9, $\lambda_{\text{sli}}$ and $\lambda_{\text{data}}$ are set to $0.05$ and $0.1$. In the final refinement optimisation step, we use $\lambda_{\text{data}}=1.0$ while keeping the same weights for the other terms. Rather than using a Chamfer loss for $\mathcal{L}_{\text{con}}$ to minimise the body-environment contact vertex distance, we use the Hausdorff measure \cite{HausdorffDist}; indeed, we observed that, with this measure, the reconstructed 3D motion is more robust to the false positive contact labels on the environment vertices.  Note that the 2D keypoints are normalised by the image size. The joint angles are defined in radian. 

\subsection{Random Sampling for Root Translation and Orientation}
\label{ssec:rand_samp}
In the pose manifold sampling-based optimisation stage, we generate candidate pose samples in the learned manifold space as described in the main paper (see Fig.\,2-(b) for its schematic visualisation). For the root translation and orientation, we generate random samples around the initial translation $\btau_{\text{opt}}$ and $\bphi_{\text{opt}}$ since they have only 3 DoF for each. Specifically, we generate samples by adding the randomly generated offsets $\Delta\btau = \psi \varphi_{\btau}$ and $\Delta\text{opt}= \psi \varphi_{\bphi}$ to $\btau_{\text{opt}} $ and $\bphi_{\text{opt}}$, respectively; $\psi$ is initialised to $1.0$, and incremented by 1 when the solution is not found due to the hard collision constraint; $\varphi_{\btau} \in [-0.03,0.03]^{3}$ and $\varphi_{\bphi} \in [-0.01,0.01]^{3}$ are the values generated uniformly at random. The range of $\varphi_{\bphi}$ is kept small since even a small change of the root orientation largely modifies the 3D joint positions.

\subsection{Random Sampling in the kinematic pose space for the ablation}
\label{ssec:rand_samp_pose}
Here, we elaborate on the details of the random sampling strategy used  for the ablative study: ``\textit{our manifold sampling strategy vs. na\"ive random sampling with a uniform distribution in a kinematic skeleton frame}'' in Sec.~5.1 in the main paper. 
 Specifically, for the na\"ive random sampling, we use the random sampling for the pose parameter $\btheta_{\text{opt}}\in \mathbb{R}^{3K}$ similar to the method explained in Sec.~\ref{ssec:rand_samp}: the randomly generated offsets $\Delta\btheta=\psi \varphi_{\btheta}$ are added to $\btheta_{\text{opt}}$ to generate the pose samples; $\varphi_{\btheta}\in [-0.26,0.26]^{3K}$ are the values that are uniformly generated at random.%

\section{Dataset and Training Details}
\label{sec:data_train}
As elaborated in Sec.\,4 in the main paper, we use GTA-IM \cite{caoHMP2020} and PROX dataset \cite{PROX:2019} for our network training. We first pretrain our networks on the whole GTA-IM dataset using the image sequences and our body contact annotations. Lastly, we train our networks on PROX dataset with the environment contact labels obtained by us (see Sec.\,4). The script to obtain the contact labels from PROX dataset and the annotated contact labels on GTA-IM dataset will be released for the future comparisons. For the evaluations, we extract test sequences from PROX and GPA \cite{gpa2019,cdg2020} datasets. The test sequence IDs are listed in the file \texttt{test_seq_IDs.txt} in our supplement. To report the 3D per-vertex errors on GPA dataset in our main paper, we fit the SMPL-X human mesh model onto the ground-truth 3D joint keypoints. The script for this operation will also be released. In addition to the recordings in indoor scenes, PROX dataset also provides the studio recordings of the accurate 3D human shape and pose for quantitative comparison purposes. However, these recordings provide the non-contiguous sequences, hence not suitable for the evaluations of our method that requires the contiguous image sequences as one of the inputs. Therefore, we mainly report the quantitative results on GPA dataset \cite{gpa2019,cdg2020}, and qualitative results on PROX dataset.
During the training, the ground-truth scene contact vertex information is once converted into the frustum voxel grid representations as elaborated on Sec.\,\ref{ssec:frustum_conversion}. We further apply 3D Gaussian filtering to obtain the smoothed contact label signal, which helps to stabilise the network training. 

\section{Network Details} 
\label{sec:net_arch}
\begin{figure*}[ht!]
	\begin{center}
		\includegraphics[width=\linewidth ]{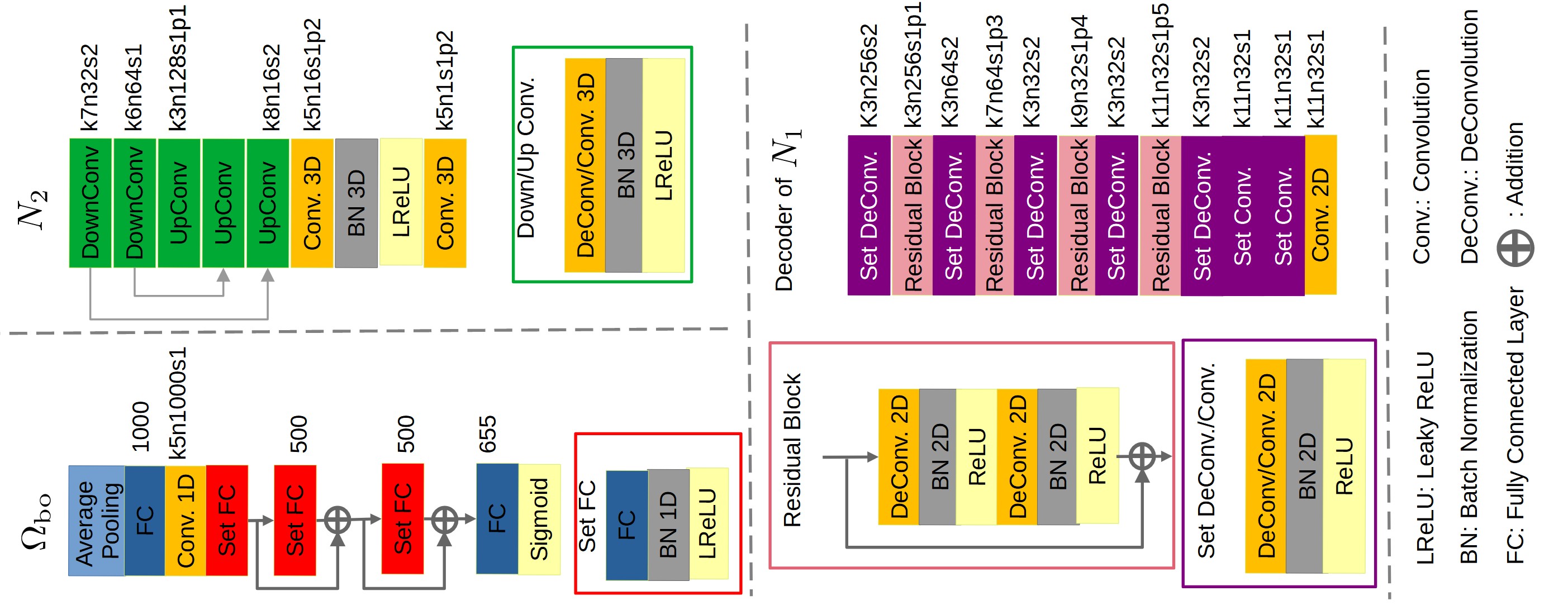}
	\end{center}
	\caption
	{ \label{fig:net_arch} The detailed network architectures for $N_{2}$, $\Omega_{\text{bo}}$ and the decoder of $N_{1}$. The numbers next to the fully-connected layers represent the output dimensionality. The numbers next to the convolution layers represent kernel size (`k'), number of kernels (`n'), size of sliding (`s') and padding size (`p'). Note that when the padding size is not shown, no padding is applied at the convolution layer.
	} 
\end{figure*}

We elaborate here on the network architectures in the dense contact estimation stage. Networks $N_{1}$ and $N_{3}$ consist of 2D-convolution-based encoder and decoder architectures. We employ Resnet-18 \cite{resnet} for the encoder of $N_{1}$ without the last two layers, i.e., a fully-connected layer and an average-pooling layer. We employ U-Net \cite{ronneberger2015u}-based architecture for $N_{3}$ with 2 sets of down-convolution and up-convolutions blocks. Network $\Omega_{\text{en}}$ consists of 3 fully-connected layers with LeakyReLU \cite{maas2013rectifier} activation function. At the output layer, we use a sigmoid function instead of LeakyReLU. For the details of $N_{2}$, $\Omega_{\text{bo}}$ and the decoder of $N_{1}$, please see Figure~\ref{fig:net_arch}.

\noindent\textbf{Why This Architecture Design?} Here, we discuss the architecture design choice for the environment contact estimation networks. 
Instead of the pixel-aligned network $\Omega_{\text{en}}$, a 3D-convolution-based network can also be applied to obtain the voxel grid that contains per-voxel contact labels of the 3D scene. However, we observed that the 3D-convolution-based classifier network suffers from the underfitting issue during the training due to the very small number of ground-truth positive contact labels over the total number of voxels in the grid. 
With the pixel-aligned implicit field, we can adjust these \textit{unbalanced} positive and negative contact labels by manipulating the sampling points in the 3D scene which we can freely control. 
Also, unlike the original work \cite{saito2019pifu} that provides the scalar value as a depth query, we provide a one-hot vector as a depth query to $\Omega_{\text{en}}$: we observed that it significantly reduces the loss value during the training compared to providing the scalar depth queries. 

\noindent\textbf{Loss Function Design (Eq.\,2 in the main paper).}
Binary GT environment contact label (‘1’: contact, ‘0’: no contact) is a very sparse signal, i.e., only a small number of voxels (\textasciitilde$0.01\%$) contain `1'. This reduces the network training stability.
We observe that smoothing the environment contact labels mitigates the unbalance and enhances the training stability. 
Hence, with smoothing, L2-loss (not BCE) for the environment contact estimation is used. Contact labels for body are more balanced compared to the environment contacts. Therefore, we do not smooth them and use BCE loss.

\section{Further Evaluations and Ablations} 
\label{sec:eva_and_abl}
In addition to the ablation studies and evaluations reported in our main paper, we further assess the performance of the dense contact estimation network (Sec.~\ref{ssec:contact_eval}) and the sliding loss term $\mathcal{L}_{\text{sli}}$ (Eq.\,10) introduced in the main paper (Sec.~\ref{ssec:sli_abl}). Lastly, we explain the setup of the qualitative results in our video (Sec.~\ref{ssec:qual_video}).

\subsection{Contact Classifications} 
\label{ssec:contact_eval}
As HULC is the first method estimating contact labels on dense body and environment surfaces from monocular RGB and point cloud input, there are no other existing works that estimate the same outputs. 
Nonetheless, we report the performance on the GPA dataset for completeness and future reference. 
The precision, recall and accuracy of the body surface contact estimation are 0.22, 0.41 and 0.91, respectively. 
For the environment surface contact estimation, 0.045, 0.18 and 0.96, respectively. 
Note that these classification tasks are highly challenging, especially since the environment point cloud contains several thousands of vertices to be classified. Furthermore, GPA dataset sequences are not included in the training dataset for the contact estimation networks (see Sec.\ref{sec:data_train} for the training/test splits).
Although it is conceivable that the reported numbers can be further improved, our framework largely benefits from the estimated contact labels and significantly reduces the 3D localisation errors as reported in Tables 2 and 3 in the main paper. 

\subsection{More Ablations for the optimisations}
\label{ssec:sli_abl}
\renewcommand{\thetable}{D\arabic{table}}
\begin{table} [t!] 
    \caption{\label{tab:abl_sliding} Ablation study for the sliding loss term $\mathcal{L}_{\text{sli}}$.} 
      \centering 
      \scalebox{0.85}{ 
      \begin{tabular} { ccc  }\toprule
	   & MPJPE [mm] $\downarrow$ & sliding error [mm] $\downarrow$ \\   \midrule
     Ours  & \bf{217.9} & \textbf{16.0} \\
     Ours (w/o $\mathcal{L}_{\text{sli}}$)  & 220.2  & 18.5 \\\bottomrule
\end{tabular}} 
\end{table}
In the main paper, we performed substantial ablation studies; \textcolor{red}{\textbf{(i)}} 3D errors and physical plausibility measurement with the variants of our method (``Ours (w/o S)'', ``Ours (w/o R)'' and ``Ours (w/o SR)'') in Tables 2 and 4, \textcolor{red}{\textbf{(ii)}} with and w/o (denoted as ``Baseline'') contact loss term $\mathcal{L}_{\text{con}}$ (Eq.\,6) in Table 3, \textcolor{red}{\textbf{(iii)}} experiments with different number of samples in the sampling-based optimisation stage (Fig. 3-a), \textcolor{red}{\textbf{(iv)}} experiments with different number of iterations in the sampling-based optimisation (Fig. 3-b), and \textcolor{red}{\textbf{(v)}} with and w/o the confidence merging strategy (Eq.\,7) in Sec.~5.1. 

For the completion, we report the ablative study for the sliding loss term $\mathcal{L}_{\text{sli}}$ (Eq.\,9) used in our optimisations. In Table \ref{tab:abl_sliding}, we report MPJPE and sliding error $e_{\text{sli.}}$ measured in a world frame for our full framework (``Ours'') and our framework w/o the sliding loss term (``Ours (w/o $\mathcal{L}_{\text{sli}}$)''). The sliding error $e_{\text{sli.}}$ is measured by computing the average of the drift of the contact vertex on the human surface, based on the assumption that contact positions in the scene are not moving (\textit{i.e.,} zero velocity). This is a reasonable assumption since most of the contact positions in daily life in a static scene are static contacts, which is also the case with our evaluation dataset; GPA dataset \cite{gpa2019,cdg2020}.

With the sliding loss term, our framework reduces the sliding error by ${\sim} 14\%$ compared to w/o $\mathcal{L}_{\text{sli}}$. Notably, integration of $\mathcal{L}_{\text{sli}}$ reduces the 3D joint error (MPJPE) by $1\%$ as well. Note that the ablation studies of the loss terms (\textit{e.g.,} 2D reprojection and temporal smoothness terms) other than $\mathcal{L}_{\text{sli}}$ and $\mathcal{L}_{\text{con}}$ (Tab. 3 in the main paper) are not of interest as those are already widely used in many wroks in video-based monocular 3D human MoCap and their significance is already well known.

\begin{figure*}[t!]
\centerline{\includegraphics[width=1\linewidth]{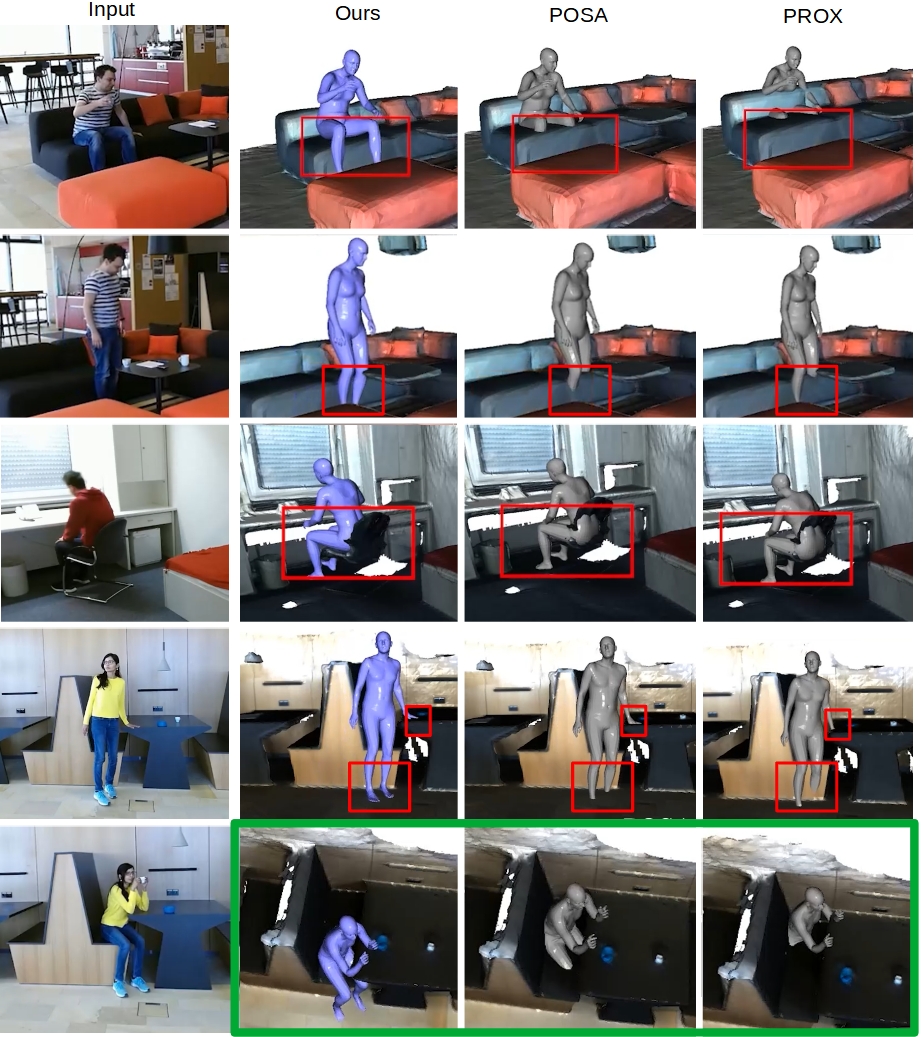}}
\caption{Qualitative comparison of our HULC \textit{vs} existing scene-aware RGB-based methods on PROX \cite{PROX:2019} dataset. Our method shows significantly mitigated collisions thanks to our novel sampling-based optimisation, which handles the severe body-environment penetrations in a hard manner (red rectangles). We also show the results from a top view (green rectangle). Thanks to the contact-based optimisation using the estimated dense contacts on the body surfaces and the environment, our estimated 3D global root positions are significantly more accurate 
compared to the previous methods. 
}
\label{fig:extra_qual} 
\end{figure*} 

\subsection{Qualitative Comparisons}
\label{ssec:qual_video}
In our video, we compare our HULC with the SOTA methods PROX\cite{PROX:2019} and POSA\cite{Hassan:CVPR:2021} from the RGB-based algorithm class. 
From the RGB-D based algorithm class, we choose PROX-D \cite{PROX:2019} and LEMO\cite{Zhang:ICCV:2021}. 
For a fair comparison, Gaussian smoothing is applied to all those related methods. 
In Fig.~\ref{fig:extra_qual}, we show comparisons of our method with other RGB-based scene-aware methods POSA\cite{Hassan:CVPR:2021} and PROX \cite{PROX:2019}. 
Our method shows physically more plausible interactions with the environment than the others. 
We also visualise the result from a bird's eye view to show the significance of the contact-based optimisation, which contributes to substantially more accurate global translation estimation than other related methods (green rectangle). 
\clearpage 
\end{document}